\documentclass[runningheads]{llncs}
\usepackage[T1]{fontenc}
\usepackage{booktabs}
\usepackage{caption} 
\usepackage{amsmath}
\usepackage{amssymb}
\usepackage{mathtools}
\usepackage{algorithm}
\usepackage{algpseudocode} 
\usepackage{multirow}
\usepackage{float}
\usepackage[table]{xcolor}
\definecolor{Best}{HTML}{DFF4E5}    
\definecolor{Second}{HTML}{FFF5D6}  
\definecolor{BaselineRow}{HTML}{F2F2F2}  
\newcommand{\baseline}[1]{\cellcolor{BaselineRow}{#1}}

\usepackage{graphicx,verbatim}
\usepackage{textcomp}
\usepackage{color}
\usepackage{xcolor}
\begin{document}
\title{Stabilizing Temporal Inference Dynamics for Online Surgical Phase Recognition}
\titlerunning{Stabilizing Temporal Inference Dynamics for Online SPR}

\author{
Yang Liu\inst{1}\textsuperscript{*} \and
Ning Zhu\inst{2}\textsuperscript{*} \and
Jingjing Peng\inst{1} \and
Xiwu Chen\inst{3} \and
Alejandro Granados\inst{1} \and
Guotai Wang\inst{2} \and
Sebastien Ourselin\inst{1}
}

\authorrunning{Y. Liu et al.}

\institute{
King's College London, London, UK\\
\email{yang.9.liu@kcl.ac.uk}
\and
University of Electronic Science and Technology of China, Chengdu, China\\
\and
Mach Drive\\
[1mm]
\textsuperscript{*}Equal contribution.
}

\maketitle

\begin{abstract}

Online Surgical Phase Recognition (SPR) models can reach high frame-wise accuracy, yet their predictions often lack temporal stability, fragmenting workflow understanding and reducing the reliability of downstream assistance. We show that this instability is not random noise but arises from two mechanisms: early misclassifications corrupt temporal feature states and propagate forward to form error cascades, and phase transitions follow evidence-accumulation dynamics whereas most online SPR systems rely on memoryless frame-wise decisions, making them sensitive to transient confidence fluctuations.
We propose a unified Train–Inference–Evaluation framework that explicitly stabilizes temporal inference dynamics using model-agnostic, plug-and-play components. For training, the Temporal Error-Cascade (TEC) loss suppresses error onset and mitigates forward error propagation by stabilizing temporal feature evolution. For inference, the Evidence-Gated Transition Predictor (EGTP) enforces evidence-driven state transitions, allowing phase changes only when accumulated evidence exceeds a confidence boundary. For evaluation, we introduce the Temporal Fragmentation Index (TFI), a reliability-aware metric that quantifies instability-induced temporal disagreement beyond conventional frame-wise and token-based measures.
Experiments on Cholec80 and AutoLaparo across three representative backbones show that the proposed framework substantially improves temporal stability and reduces prediction fragmentation, while maintaining or modestly improving frame-wise performance.


\end{abstract}

\section{Introduction}
\label{sec:intro}
Surgical Phase Recognition (SPR) aims to automatically allocate the surgical workflow phase of each frame in operation videos, serving as a fundamental component for intelligent intraoperative monitoring~\cite{Cleary2004OR2020,Franke2018IntelligentOR}, workflow assessment~\cite{Dias2019MachineLearningCompetence,Kowalewski2019SensorBasedWorkflow}, and automated documentation and alerts~\cite{Quellec2015RealTimeTaks}. Accurate and reliable SPR provides important support for post-operative quality assessment and surgical training. With the development of deep learning, significant progress has been made in improving SPR accuracy~\cite{Garrow2021Machine}. Early approaches relied on LSTM~\cite{Jin2021Temporal,Gao2020Automatic,Jin2018SVRCNET} and TCN~\cite{Czempiel2020} architectures to model temporal dependencies, while Transformer-based methods~\cite{Girdhar2021AnticipativeVideoTransformer,Yang2024Surgformer,Liu2025LoViT,Liu2023SKiT} further pushed performance by capturing long-range contextual information and quickly became the dominant paradigm. More recently, state space models such as Mamba~\cite{Gu2024Mamba} have been introduced to enable efficient long-sequence modeling with linear computational complexity~\cite{Wu2025holistic}.

However, existing methods mainly optimize accuracy while neglecting temporal stability~\cite{Chen2025SurgPLAN}. Temporal stability refers to the structural consistency of phase trajectories over time, avoiding short-term oscillations or unnecessary switches when no true transition occurs~\cite{cao2023intelligent}. In clinical scenarios, such unstable predictions fragment workflow understanding and reduce the reliability of downstream assistance. We argue that temporal instability arises from two factors. First, early misclassifications corrupt temporal feature representations and propagate forward through context-dependent models, destabilizing feature evolution dynamics and forming error cascades that compromise subsequent predictions. Second, true surgical phase transitions are inherently evidence-accumulation processes. We do not claim that all instability originates from transition uncertainty alone. Ambiguous phase definitions, annotation subjectivity, and events occurring outside the camera field of view may also contribute to unstable predictions. Our focus is on the complementary and model-agnostic problem of reducing avoidable decision oscillations caused by transient confidence fluctuations. Although modern SPR models encode temporal context in feature representations, their decision policies remain effectively memoryless given the current feature state: predictions are determined solely by instantaneous classifier outputs without explicitly modeling the causal evolution and stability of decision states. This mismatch between evidence-accumulation transition dynamics and memoryless frame-wise decisions makes predictions vulnerable to transient confidence fluctuations, resulting in fragmented phase trajectories. Despite recent attempts to mitigate instability through proposal-based rectification~\cite{Chen2025SurgPLAN}, learned refinement stages~\cite{Yi2023NotEnd}, historical feature fusion~\cite{chen2025DSTED} or explicit transition modeling~\cite{ding2025neuralfinitestatemachinessurgical}, these approaches are often tightly coupled with specific architectures, require extra training, or impose strong procedural priors, limiting their generality. More importantly, stability is typically treated as a by-product of accuracy improvement rather than an explicit objective for quantified evaluation, although higher accuracy does not guarantee coherent temporal dynamics. Therefore, achieving stable predictions in a model-agnostic manner remains an open challenge.

Moreover, quantifying temporal stability is equally critical for reliable SPR~\cite{dergachyovaAutomaticDatadrivenRealtime2016}. While \textit{Edit Score}~\cite{lea2016segmental} is widely adopted from action segmentation~\cite{Lea2016Temporal} to evaluate temporal consistency~\cite{Lea2017Temporal,Li2023MSTCN}, it operates on compressed segment tokens and disregards temporal structure within segments. As a result, predictions with drastically different stability and accuracy may receive identical Edit Scores, since instability-induced disagreement within segments is ignored. This limitation prevents existing metrics from correctly assessing prediction reliability and resolving clinically relevant trade-offs between accuracy and stability. Therefore, a dedicated metric that directly captures temporal fragmentation and instability-induced disagreement is essential for SPR temporal reliability evaluation.

In this paper, we present the first unified Train-Inference-Evaluation framework that stabilizes temporal inference dynamics at both the feature and decision level in a model-agnostic and plug-and-play manner. (1) \textbf{Training}: We introduce the Temporal Error-Cascade (TEC) loss, which suppresses error onset and prevents error propagation by stabilizing temporal feature evolution. (2) \textbf{Inference}: We propose the Evidence-Gated Transition Predictor (EGTP), a causal module that aligns phase transitions with accumulated evidence, enforcing temporally consistent decision dynamics. (3) \textbf{Evaluation}: We present the Temporal Fragmentation Index (TFI), a reliability-aware metric that quantifies instability-induced temporal disagreement and resolves ambiguities in existing metrics. (4) Across three representative SPR backbones, our framework yields consistent accuracy gains alongside \textbf{order-of-magnitude reductions} in temporal fragmentation.

\section{Methodology}
\label{sec:method}
\begin{figure}[t]
  \centering
  \includegraphics[width=\linewidth]{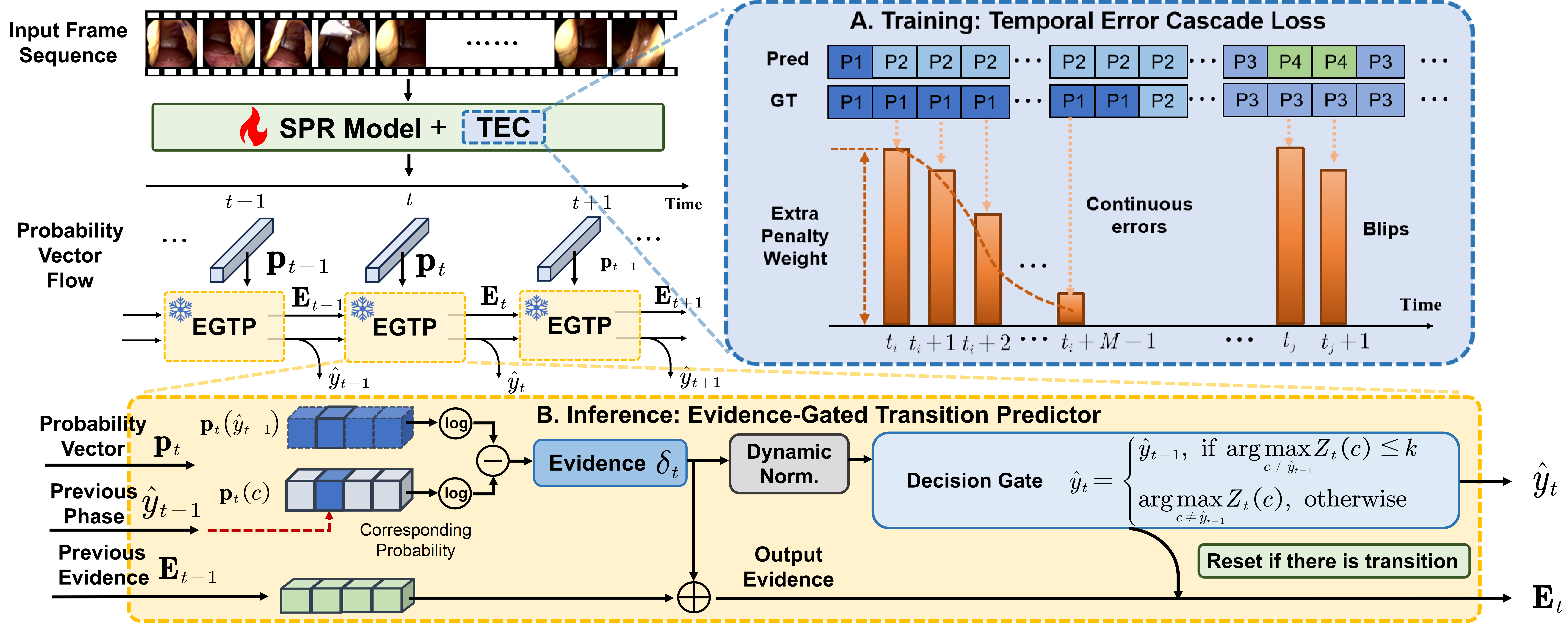}
  \caption{Overview of proposed TEC for training and EGTP for inference.}
  \label{fig:method}
\end{figure}

Given an input surgical video stream $\mathbf{X}=\{x_t\}_{t=1}^{T}$, online SPR predicts a phase sequence $\hat{\mathbf{y}}=\{\hat{y}_t\}_{t=1}^{T}$ causally, where each prediction $\hat{y}_t \in \{1,\dots,N\}$ depends only on past and current observations $\{x_\tau\}_{\tau \le t}$ and the ground truth is $\mathbf{y}=\{y_t\}_{t=1}^{T}$. We propose a unified framework that explicitly stabilizes temporal dynamics at both the feature and decision levels. During training, we introduce the Temporal Error-Cascade (TEC) loss to mitigates forward corruption of temporal feature representations. During inference, we apply the Evidence-Gated Transition Predictor (EGTP) to enforce temporally consistent decision transitions. Finally, we evaluate prediction reliability using the proposed Temporal Fragmentation Index (TFI). An overview is shown in \textbf{Fig.~\ref{fig:method}}.

\subsection{Training: Temporal Error-Cascade Loss Function}
Standard objectives such as cross-entropy penalize misclassifications uniformly, implicitly treating errors as independent over time. In temporal models, however, an early misclassification can corrupt the temporal state, so subsequent features and predictions become conditioned on this corrupted state, leading to a forward \emph{error cascade}. To explicitly stabilize temporal feature evolution, we propose the Temporal Error-Cascade (TEC) loss as shown in Fig.~\ref{fig:method}(A), which suppresses the \emph{onset} of consecutive errors, the point at which state corruption is initiated.
We define $e_t=\mathbb{I}\!\left(\hat{y}_t \neq y_t\right)$ and partition consecutive misclassified frames into maximal error runs $\mathcal{R}=\{[s_r,e_r]\}_{r=1}^{N_R}$ with length $L_r=e_r-s_r+1$. For each run, we reweight only its onset window of length $L_r^{(M)}=\min(L_r,M)$. If a frame $t$ lies in $[s_r,\,s_r+L_r^{(M)}-1]$, we apply a front-loaded Gaussian weight decaying with $t-s_r$ frames; otherwise the weight is $1$:
\begin{equation}
w_t = 1 + \alpha \sum_{r=1}^{N_R} \mathbb{I}\!\big[t \in [s_r,\, s_r + L_r^{(M)} - 1]\big]
\exp\!\left( -\frac{(t - s_r)^2}{2\sigma^2} \right).
\end{equation}
Thus, correctly classified frames ($e_t=0$) and frames beyond the first $M$ positions of an error run retain the standard cross-entropy weight. The TEC loss is defined as the weighted cross-entropy:
\begin{equation}
\mathcal{L}_{TEC}=\frac{1}{T}\sum_{t=1}^{T}w_t \cdot \ell_{CE}.
\end{equation}
where $\ell_{CE}$ denotes the per-frame cross-entropy loss.
By concentrating the penalty on error onset, the point at which temporal state corruption is initiated, TEC encourages rapid recovery before corrupted representations propagate forward, thereby stabilizing temporal feature dynamics in a model-agnostic manner.

\subsection{Inference: Evidence-Gated Transition Predictor}
Although modern SPR backbones encode temporal context in feature representations, predictions are typically produced by frame-wise $\arg\max$ decisions, which constitute a memoryless decision policy. As a result, predictions near decision boundaries are sensitive to transient confidence fluctuations, causing unstable phase switching even when no true transition occurs.
To explicitly stabilize temporal decision dynamics, we propose Evidence-Gated Transition Predictor (EGTP), a causal and plug-and-play \emph{state transition policy} that models phase changes as evidence accumulation.  EGTP maintains the current state $\hat{y}_{t-1}$ and permits a transition only when sufficient accumulated evidence supports an alternative phase.
At time $t$, the backbone outputs class probabilities $\mathbf{p}_t \in [0,1]^N$. For each candidate phase $c \neq \hat{y}_{t-1}$, EGTP computes instantaneous log-likelihood evidence against the current state:
\begin{equation}
\delta_t(c) =
\log \mathbf{p}_t(c)
-
\log \mathbf{p}_t(\hat{y}_{t-1}).
\end{equation}
Evidence is accumulated with a one-sided hysteresis update:
\begin{equation}
\mathbf{E}_t(c)
=
\max\!\left(0,\,
\mathbf{E}_{t-1}(c) + \delta_t(c)
\right),
\end{equation}
which allows consistent evidence to build up while quickly discarding temporary reversals.
To ensure robustness across models and probability scales, accumulated evidence is normalized:
\begin{equation}
Z_t(c)
=
\frac{\mathbf{E}_t(c)}
{\sqrt{n_t(c)}\,\sigma_t(c)},
\end{equation}
where $n_t(c)$ and $\sigma_t(c)$ denote the number and running standard deviation of evidence. EGTP updates the phase using an evidence-gated transition rule:
\begin{equation}
\hat{y}_t =
\begin{cases}
\displaystyle
\arg\max_{c \neq \hat{y}_{t-1}} Z_t(c),
&
\text{if }
\max_{c \neq \hat{y}_{t-1}} Z_t(c) > k,
\\[6pt]
\hat{y}_{t-1},
&
\text{otherwise.}
\end{cases}
\end{equation}

After a transition, accumulated statistics are reset to zero to prevent immediate reversal. By replacing memoryless frame-wise decisions with causal evidence-driven state transitions, EGTP stabilizes temporal decision dynamics while remaining fully causal and model-agnostic.

\subsection{Evaluation: Temporal Fragmentation Index}
Existing SPR studies primarily evaluate models using accuracy or segment-level metrics such as \textit{Edit Score}. However, they fail to reflect temporal prediction reliability, which depends jointly on prediction correctness and temporal consistency. In clinical deployment, models often exhibit trade-offs between accuracy and stability, and conventional metrics cannot objectively determine which prediction is more reliable. Importantly, accuracy and Edit Score may assign identical scores to predictions with drastically different reliability. For example, given ground truth $\texttt{AAAAA\;BBBB\;CCC}$, predictions $\texttt{AAAAA\;AAAA\;AAA}$ (stable and more accurate) and $\texttt{CABCB\;BAAC\;AAB}$ (unstable and less accurate) achieve identical Edit Score (33.33), despite significantly different temporal stability and correctness. This occurs because Edit Score evaluates only compressed segment tokens and ignores temporal disagreement within segments. To address this, we propose TFI, a unified metric that quantifies temporal reliability via instability-induced disagreement, rather than measuring fragmentation alone. Given the predicted sequence $\hat{\mathbf{y}}=\{\hat{y}_t\}_{t=1}^{T}$, we partition it into $K$ consecutive constant segments $\{[s_i,e_i)\}_{i=1}^{K}$ corresponding to maximal runs of consistent predictions. Let $G$ denote the number of ground truth segments and per-video TFI is defined as
\begin{equation}
\mathrm{TFI}(\hat{\mathbf{y}},\mathbf{y})
=
\frac{1}{G}
\sum_{i=1}^{K}
\Big(\frac{1}{e_i - s_i}
\sum_{t=s_i}^{e_i-1}
\mathbb{I}\!\left(\hat{y}_t \neq y_t\right)\Big),
\end{equation}
which measures the average instability-induced disagreement ratio within each prediction run. The dataset-level TFI is computed by averaging per-video scores across all videos.
Unlike accuracy (error frequency) and Edit Score (segment ordering), TFI captures how errors are distributed over time. Fragmented predictions create more runs and disagreement, increasing TFI even when conventional metrics are unchanged. Thus, TFI provides a principled reliability measure that distinguishes stable, reliable trajectories from fragmented, unreliable ones.

\section{Experiments}
\label{sec:exp}

\subsubsection{Experimental Setup.}

We evaluate our framework on two public benchmarks, Cholec80~\cite{Twinanda2017Cholec80} and AutoLaparo~\cite{Wang2022AutoLaparo}. Cholec80 consists of 80 cholecystectomy videos annotated with 7 surgical phases, split into 40 training and 40 testing cases. AutoLaparo contains 21 laparoscopic hysterectomy videos with 7 phases, divided into 10 training, 4 validation, and 7 testing videos. Following prior work~\cite{Jin2021Temporal,Liu2025LoViT}, we report video-level Accuracy and phase-level Precision, Recall, and Jaccard. To explicitly evaluate temporal coherence, we additionally report Edit Score and the proposed TFI. TEC and EGTP are integrated into three representative online SPR backbones: Trans-SVNet~\cite{Gao2021Trans-SVNet}, SKiT~\cite{Liu2023SKiT}, and Surgformer~\cite{Yang2024Surgformer} based on their official implementations. For TEC, we set $\sigma=1.5$ and $M=8$, with $\alpha=7$ for Cholec80 and $\alpha=19$ for AutoLaparo. For EGTP, the confidence threshold $k$ is set to $0.4$ and $0.8$ on the two datasets, respectively.

\begin{table*}[t]
\centering
\caption{Overall comparison of SPR backbones with proposed TEC and EGTP.}
\label{tab:overall_comparison}
\resizebox{\textwidth}{!}{
\begin{tabular}{c|cccccc|cccccc}
\hline
\multirow{2}{*}{\textbf{Methods}} 
& \multicolumn{6}{c|}{\textbf{Cholec80 Dataset}} 
& \multicolumn{6}{c}{\textbf{AutoLaparo Dataset}} \\

& \textbf{Acc} $\uparrow$ & \textbf{Pre} $\uparrow$ & \textbf{Rec} $\uparrow$ & \textbf{Jac} $\uparrow$ 
& \textbf{Edit} $\uparrow$ & \textbf{TFI} $\downarrow$
& \textbf{Acc} $\uparrow$ & \textbf{Pre} $\uparrow$ & \textbf{Rec} $\uparrow$ & \textbf{Jac} $\uparrow$ 
& \textbf{Edit} $\uparrow$ & \textbf{TFI} $\downarrow$ \\
\hline

Trans-SVNet~\cite{Gao2021Trans-SVNet}
& 88.61 & 84.08 & 83.23 & 71.74 & 12.54 & 6.42 
& 80.16 & 68.73 & 62.65 & 51.22 & 2.94 & 18.80 \\

Trans-SVNet~\cite{Gao2021Trans-SVNet}+TEC
& 89.01 & 83.43 & 83.94 & 71.98 & 14.29 & 5.87 
& 82.02 & 70.29 & 65.93 & 55.03 & 3.17 & 16.09 \\

Trans-SVNet~\cite{Gao2021Trans-SVNet}+EGTP
& 89.09 & \textbf{84.96} & 83.35 & \textbf{72.64} & 51.52 & 1.19 
& 80.53 & 68.00 & 62.21 & 51.19 & 39.65 & 1.19 \\

Trans-SVNet~\cite{Gao2021Trans-SVNet}+TEC+EGTP
& \textbf{89.13} & 84.34 & 83.53 & 72.56 & \textbf{54.97} & \textbf{1.16} 
& \textbf{83.79} & \textbf{71.10} & \textbf{66.96} & \textbf{56.89} & \textbf{56.60} & \textbf{0.26} \\
\hline

SKiT~\cite{Liu2023SKiT}
& 91.91 & 86.42 & 86.70 & 76.38 & 19.45 & 4.59 
& 81.59 & 77.95 & 71.90 & 61.29 & 5.11 & 13.38 \\

SKiT~\cite{Liu2023SKiT}+TEC
& 91.98 & 85.83 & \textbf{88.15} & \textbf{76.97} & 20.82 & 3.47 
& 82.24 & 74.68 & 72.65 & 61.57 & 5.71 & 10.11 \\

SKiT~\cite{Liu2023SKiT}+EGTP
& 92.21 & \textbf{86.89} & 85.53 & 76.15 & 61.95 & 0.78 
& 82.35 & 81.35 & 71.11 & 60.39 & 49.05 & 0.89 \\

SKiT~\cite{Liu2023SKiT}+TEC+EGTP
& \textbf{92.54} & 86.88 & 85.77 & 76.64 & \textbf{62.40} & \textbf{0.69} 
& \textbf{84.10} & \textbf{84.26} & \textbf{73.29} & \textbf{62.90} & \textbf{49.47} & \textbf{0.78} \\
\hline

Surgformer~\cite{Yang2024Surgformer}
& 91.22 & 86.22 & 87.97 & 76.98 & 15.77 & 5.85 
& 84.49 & 76.50 & 71.55 & 61.98 & 4.24 & 14.57 \\

Surgformer~\cite{Yang2024Surgformer}+TEC
& 91.78 & 86.39 & \textbf{89.33} & \textbf{78.45} & 16.60 & 5.54 
& \textbf{86.27} & 82.16 & \textbf{76.11} & \textbf{67.21} & 6.45 & 10.67 \\

Surgformer~\cite{Yang2024Surgformer}+EGTP
& 91.60 & 86.71 & 87.69 & 77.01 & 54.12 & 1.09 
& 85.15 & \textbf{87.18} & 71.89 & 62.77 & 43.68 & 0.96 \\

Surgformer~\cite{Yang2024Surgformer}+TEC+EGTP
& \textbf{92.20} & \textbf{86.89} & 88.70 & 78.30 & \textbf{60.53} & \textbf{0.71} 
& \textbf{86.27} & 78.18 & 73.79 & 64.62 & \textbf{61.77} & \textbf{0.48} \\
\hline
\end{tabular}
}
\end{table*}

\begin{figure}[t]
  \centering
  \includegraphics[width=\linewidth]{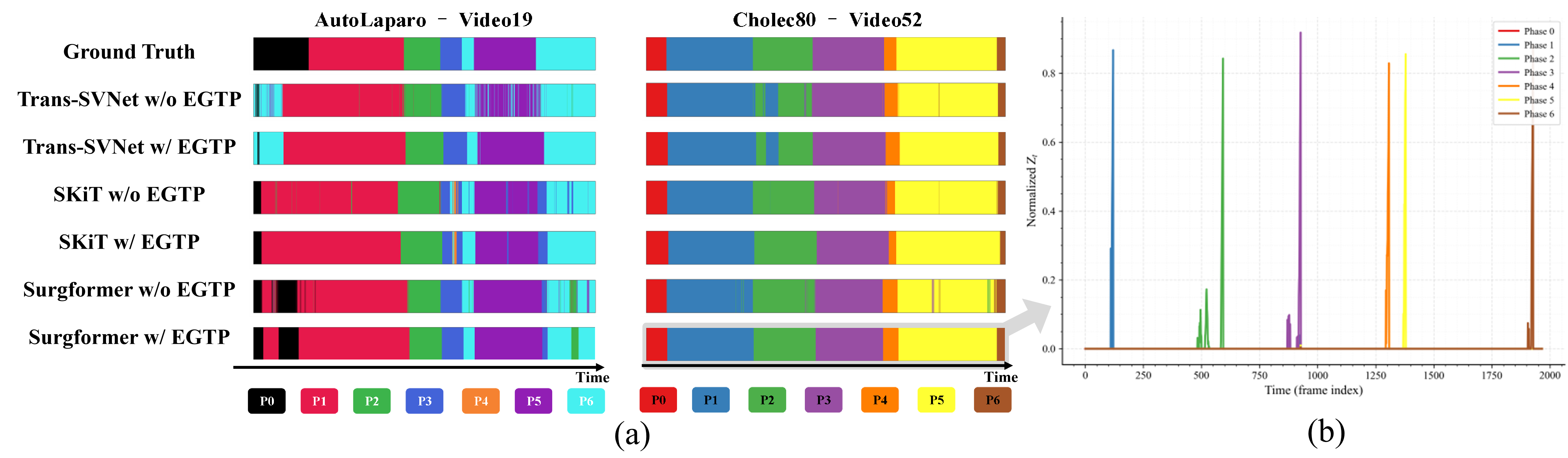}
  \caption{(a) Comparison of prediction visualization results w/ or w/o EGTP. (b) A qualitative example of temporal evolution of the normalized evidence $Z_t$.}
  \label{fig:EGHS_visual}
\end{figure}

\subsubsection{Comparative Study.}

\textbf{Table~\ref{tab:overall_comparison}} summarizes the comparative results on both datasets. Across all three backbones, both TEC and EGTP individually boost temporal stability while slightly improving accuracy. TEC mainly yields moderate but consistent reductions in TFI with stable accuracy gains (e.g., TFI reduced from 6.42 to 5.87 of Trans-SVNet on Cholec80). EGTP produces a dramatic drop in TFI, confirming its effectiveness in suppressing oscillations. On AutoLaparo of Surgformer, EGTP reduces TFI from 10.67 to 0.48. When TEC and EGTP are combined, the best overall performance is consistently achieved. On average, the joint strategy improves accuracy by $0.7\%$ on Cholec80 and $2.6\%$ on AutoLaparo compared with the original backbones, while reducing TFI by nearly \textbf{one order of magnitude}. \textbf{Fig.~\ref{fig:EGHS_visual}} further shows that EGTP can largely eliminate short-term flickers and rapid phase switching and the normalized evidence $Z_t$ reveals consistent accumulation before phase transitions, reducing spurious short-lived switches.

\begin{table*}[t]
\centering
\scriptsize
\caption{Ablation study of TEC under different $\alpha$ settings.}
\label{tab:tera_effectiveness_all}
\renewcommand{\arraystretch}{1.1}
\setlength{\tabcolsep}{4pt}
\resizebox{\textwidth}{!}{
\begin{tabular}{c|c|ccc|ccc|ccc}
\hline
\multirow{2}{*}{\textbf{Dataset}} & \multirow{2}{*}{\textbf{Setting}} 
& \multicolumn{3}{c|}{\textbf{Trans-SVNet~\cite{Gao2021Trans-SVNet}}} 
& \multicolumn{3}{c|}{\textbf{SKiT~\cite{Liu2023SKiT}}} 
& \multicolumn{3}{c}{\textbf{Surgformer~\cite{Yang2024Surgformer}}} \\
\cline{3-11}
& 
& Acc$\uparrow$ & Edit$\uparrow$ & TFI$\downarrow$
& Acc$\uparrow$ & Edit$\uparrow$ & TFI$\downarrow$
& Acc$\uparrow$ & Edit$\uparrow$ & TFI$\downarrow$ \\
\hline

\multirow{6}{*}{\textbf{Cholec80}}
& \baseline{w/o TEC}
& \baseline{88.61} & \baseline{12.54} & \baseline{6.42} 
& \baseline{91.91} & \baseline{19.45} & \baseline{4.59}
& \baseline{91.22} & \baseline{15.77} & \baseline{5.85} \\

& $\alpha=3$ 
& \underline{88.82} & \textbf{15.85} & \textbf{4.74} 
& \underline{92.14} & 21.14 & \underline{3.20} 
& 91.94 & 15.12 & 5.61 \\

& $\alpha=7$ 
& \textbf{89.01} & \underline{14.29} & 5.87 
& 91.98 & 20.82 & 3.47 
& 91.78 & 16.60 & 5.54 \\

& $\alpha=11$ 
& 88.62 & 13.84 & \underline{5.43} 
& 91.91 & \underline{22.48} & 3.36 
& 92.10 & \underline{17.92} & 5.37 \\

& $\alpha=15$ 
& 88.55 & 13.36 & 6.06 
& \textbf{92.18} & 20.92 & 3.33 
& \textbf{92.25} & \textbf{18.00} & \textbf{4.84} \\

& $\alpha=19$ 
& 88.59 & 14.01 & 5.45 
& 92.09 & \textbf{23.43} & \textbf{3.15} 
& \underline{92.20} & 17.08 & \underline{5.01} \\

\hline

\multirow{6}{*}{\textbf{AutoLaparo}}
& \baseline{w/o TEC}
& \baseline{80.16} & \baseline{2.94} & \baseline{18.80} 
& \baseline{81.59} & \baseline{5.11} & \baseline{13.38} 
& \baseline{84.49} & \baseline{4.24} & \baseline{14.57} \\

& $\alpha=3$ 
& 81.74 & 3.34 & 17.09 
& 82.21 & 5.90 & 9.60
& 84.16 & 5.20 & 11.67 \\

& $\alpha=7$ 
& 81.02 & \textbf{4.07} & 15.65
& \textbf{82.29} & \underline{6.79} & 8.86 
& 84.82 & \underline{5.25} & \underline{11.04} \\

& $\alpha=11$ 
& 81.00 & \underline{3.46} & 17.98 
& 82.21 & 6.67 & \underline{8.57} 
& 83.81 & 4.61 & 12.32 \\

& $\alpha=15$ 
& \textbf{82.09} & 3.20 & \textbf{16.10}
& 81.85 & \textbf{7.77} & \textbf{7.78} 
& \underline{85.68} & 4.93 & 13.04 \\

& $\alpha=19$ 
& \underline{82.02} & 3.17 & \underline{16.09} 
& \underline{82.24} & 5.71 & 10.11 
& \textbf{86.27} & \textbf{6.45} & \textbf{10.67} \\

\hline
\end{tabular}
}
\end{table*}

\begin{figure}[t]
  \centering
  \includegraphics[width=\linewidth]{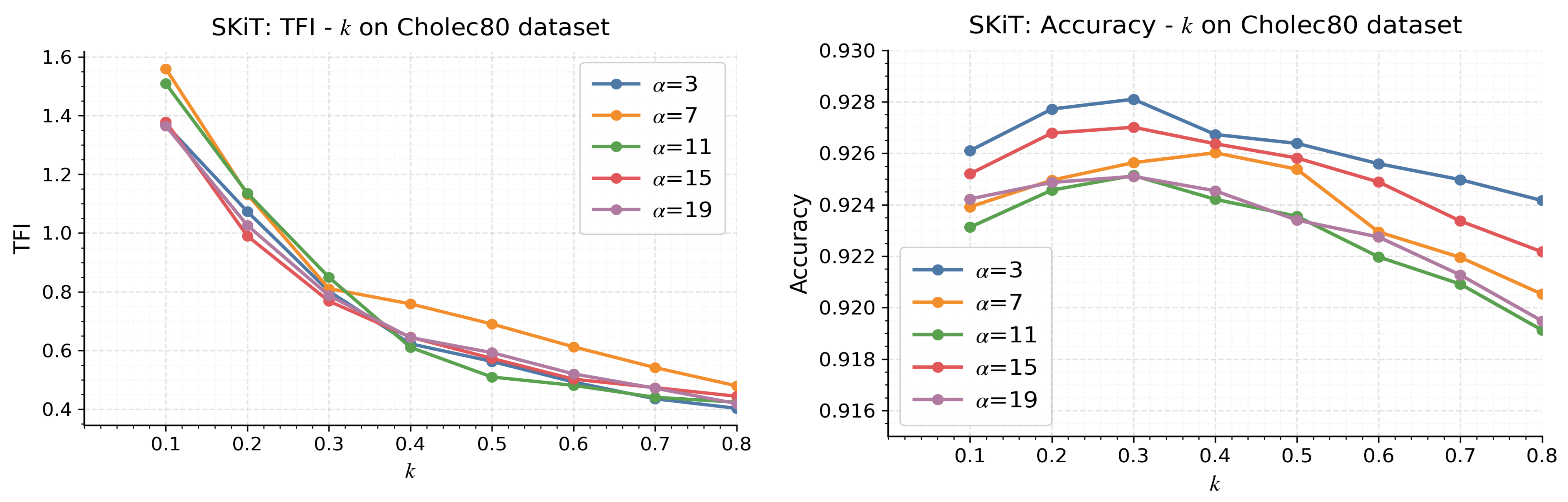}
  \caption{Relationship between accuracy/stability versus $k$ of SKiT on Cholec80.}
  \label{fig:tsi_acc}
\end{figure}

\subsubsection{Ablation Study of TEC.}
To analyze the robustness of TEC, we vary the penalty strength $\alpha \in \{3,7,11,15,19\}$. \textbf{Table~\ref{tab:tera_effectiveness_all}} shows that for nearly all backbones and datasets, introducing TEC leads to a clear TFI reduction with maintained or higher accuracy. For example, Trans-SVNet on Cholec80 reduces TFI from 6.42 ($\alpha=0$) to 4.74 ($\alpha=3$) with accuracy increasing from 88.61 to 88.82, and SKiT on AutoLaparo reduces TFI from 13.38 to 8.57 ($\alpha=11$) while also improving accuracy. Notably, Edit Score and TFI do not always follow identical trends. For instance, by comparing $\alpha=3$ with $\alpha=11$ of Trans-SVNet on AutoLaparo dataset, Edit Score suggests $\alpha=11$ is more stable (3.34 vs. 3.46), whereas TFI indicates $\alpha=3$ yields lower instability (17.09 vs. 17.98). This discrepancy highlights that the difference between compressed-level metric Edit Score and TFI that is dedicated for temporal consistency measurement.

\begin{table}[t]
\centering
\scriptsize
\caption{Comparison between EGTP and E-Const ($\alpha=19$).}
\label{tab:dynamic_norm}
\renewcommand{\arraystretch}{1.1}
\setlength{\tabcolsep}{3pt}
\resizebox{\linewidth}{!}{
\begin{tabular}{c c c c c c c c}
\hline
Method & Model & Data & Best Th. & Acc@Best Th & TFI@Best Th & Acc @Avg Th. & TFI@Avg Th. \\
\hline
\multirow[c]{4}{*}{E-Const}
        & \multirow[c]{2}{*}{Trans-SVNet~\cite{Gao2021Trans-SVNet}} & Auto. & 15   & 83.66 & 0.47 & 83.65 & 0.53 \\
        &  & Cho.   & 0.2  & 88.51 & 3.05 & 87.10 & 0.43 \\
        & \multirow[c]{2}{*}{SKiT~\cite{Liu2023SKiT}}       & Auto. & 25   & 84.10 & 0.69 & 83.50 & 1.41 \\
        &          & Cho.   & 5    & 92.27 & 0.76 & 92.15 & 0.55 \\
\hline
\multirow[c]{4}{*}{EGTP}
        & \multirow[c]{2}{*}{Trans-SVNet~\cite{Gao2021Trans-SVNet}} & Auto. & 0.6  & 84.09 & 1.08 & 84.05 & 0.92 \\
        &  & Cho.   & 0.3  & 88.71 & 1.66 & 88.19 & 0.76 \\
        & \multirow[c]{2}{*}{SKiT~\cite{Liu2023SKiT}}       & Auto. & 0.8  & 84.10 & 0.78 & 84.03 & 1.24 \\
        &        & Cho.   & 0.3  & 92.51 & 1.03 & 92.34 & 0.59 \\
\hline
\end{tabular}
}
\end{table}

\subsubsection{Ablation Study of EGTP.}

We further analyze the effect of the threshold $k$ and the role of dynamic normalization in EGTP. As illustrated in \textbf{Fig.~\ref{fig:tsi_acc}}, increasing $k$ leads to progressively lower TFI, indicating more stable predictions, while accuracy first improves and then slightly decreases. This reflects the intended stability–accuracy trade-off of EGTP: a larger $k$ enforces stronger accumulated evidence before switching to suppress transient oscillations. However, overly large $k$ may delay or miss genuine transitions, causing accuracy degradation.

To examine whether dynamic normalization enhances cross-model robustness, we construct a variant termed \textbf{E-Const}, which removes variance normalization and applies a constant threshold directly to accumulated evidence. Experiments are conducted on Trans-SVNet~\cite{Gao2021Trans-SVNet} and SKiT~\cite{Liu2023SKiT} with $\alpha=19$. For each model–dataset pair, we report the best-performing threshold and additionally test a unified threshold obtained by averaging optimal values within each method (\textbf{Table~\ref{tab:dynamic_norm}}). Although E-Const can occasionally yield lower TFI, EGTP shows two clear advantages. First, EGTP consistently achieves higher accuracy, for example 84.09\% vs.\ 83.66\% on AutoLaparo (Trans-SVNet). Second, E-Const requires highly inconsistent optimal thresholds (ranging from 0.2 to 25). This demonstrates that dynamic normalization aligns evidence scales to improve cross-backbone robustness and reduce sensitivity to manual tuning.

\section{Conclusion}
\label{sec:conclusion}
In this work, we present the first unified framework that systematically stabilizes temporal dynamics throughout the training, inference, and evaluation stages in a model-agnostic and plug-and-play fashion. First, we propose the Temporal Error-Cascade (TEC) loss to effectively suppress the onset of errors and mitigate their forward propagation and error cascades. Second, we develop the Evidence-Gated Transition Predictor (EGTP), a novel decision policy that only permits transitions with sufficient accumulated evidence, enforcing temporally consistent decision-making. Third, we present the Temporal Fragmentation Index (TFI), a novel reliability-oriented metric that quantifies instability-induced temporal disagreement within predicted segments and addresses limitations in existing evaluation measures. Extensive experiments on three representative SPR backbones demonstrate consistent accuracy gains together with order-of-magnitude reductions in segment fragmentation, establishing a model-agnostic pathway toward stable and clinically deployable surgical phase recognition systems.

\newpage
\bibliographystyle{splncs04}
\bibliography{reference}
\end{document}